\relax
\documentclass[letterpaper]{article} 

\usepackage{aaai22}  
\usepackage{times}  
\usepackage{helvet}  
\usepackage{courier}  
\usepackage[hyphens]{url}  
\usepackage{graphicx} 
\usepackage{fixltx2e}
\usepackage{natbib}  
\usepackage{caption} 
\DeclareCaptionStyle{ruled}{labelfont=normalfont,labelsep=colon,strut=off} 
\usepackage{algorithm}
\usepackage{algorithmic}
\usepackage{amsfonts}

\usepackage{newfloat}
\usepackage{listings}
\floatstyle{ruled}
\newfloat{listing}{tb}{lst}{}
\floatname{listing}{Listing}
\newcommand{\ra}[1]{\renewcommand{\arraystretch}{#1}}

\title{UserBERT: Modeling Long- and Short-Term User Preferences 
\\via Self-Supervision}
\author {
    Tianyu Li\equalcontrib,
    Ali Cevahir\equalcontrib,
    Derek Cho,
    Hao Gong,
    DuyKhuong Nguyen,
    Bj\"orn Stenger
}
\affiliations {
    Rakuten Institute of Technology, Rakuten Group, Inc.\\
    \{derek.cho, bjorn.stenger\}@rakuten.com
}

\usepackage{arydshln}
\usepackage{multirow}
\usepackage{subcaption}
\usepackage{booktabs}

\newcommand{\ie}{\textit{i}.\textit{e}.}
\newcommand{\eg}{\textit{e}.\textit{g}.}

\begin{document}

\maketitle

\begin{abstract}
E-commerce platforms generate vast amounts of customer behavior data, such as clicks and purchases, from millions of unique users every day. However, effectively using this data for behavior understanding tasks is challenging because there are usually not enough labels to learn from all users in a supervised manner. This paper extends the BERT model to e-commerce user data for pretraining representations in a self-supervised manner. By viewing user actions in sequences as analogous to words in sentences, we extend the existing BERT model to user behavior data. Further, our model adopts a unified structure to simultaneously learn from long-term and short-term user behavior, as well as user attributes. We propose methods for the tokenization of different types of user behavior sequences, the generation of input representation vectors, and a novel pretext task to enable the pretrained model to learn from its own input, eliminating the need for labeled training data. Extensive experiments demonstrate that the learned representations result in significant improvements when transferred to three different real-world tasks, particularly compared to task-specific modeling and multi-task representation learning.

\end{abstract}
\section{Introduction}

The choice of data representation, \ie, how to extract meaningful features, has significant impact on the performance of machine learning applications \citep{bengio2013survey}. Therefore, data processing and feature engineering have been key steps in model development. 
To extend the applicability of the models, recent research on representation learning aims to discover the underlying explanatory factors hidden in raw data.
With rapid advances in this direction, we have witnessed breakthroughs in the areas of computer vision \citep{doersch2015context,razavian2014,SimoSerraICCV2015} and natural language processing (NLP) \citep{mikolov2013word2vec, pennington2014glove, Lin2017ASS}. 

Similarly, for building user-oriented industrial applications like next purchase prediction and recommendation, much effort has been spent on understanding business models and user behavior for creating useful features \citep{richardson2007predicting, covington2016youtube}. However, this is a time-consuming and application-specific process. It is also challenging to reuse these features, or to share the gained knowledge between different services and tasks.

To address the issue of isolated feature engineering and task-specific design, prior work has explored pretraining and transfer learning ideas. For example, multi-task learning (MTL) has shown promising results~\citep{Ni2018PerceiveInDepth}. However, MTL has intrinsic challenges, \eg,  deciding which tasks to learn jointly \citep{Standley2019WhichTS}, or how to weigh tasks to achieve optimal performance~\citep{Kendall2018MultitaskLU}. More importantly, the training still hinges on large amounts of well-annotated user labels.



Inspired by the BERT model, which has been immensely useful across a host of NLP tasks \citep{Devlin2019BERTPO, lan2020albert}, 
recent work proposed pretraining user representations on unlabeled behavior data in a self-supervised manner \citep{Wu2020PTUMPU, Yuan2020peterrec}.
However, prior work does not take inherent differences between different types of user behavior into account.
Our proposal, \emph{UserBERT}, simultaneously learns from three categories of user data, \ie, long-term and short-term behavior, as well as user attributes, via a unified architecture.
In this work, we consider {\it short-term behavior} as  user actions during one browsing session, including clicks, searches, and page views.
{\it Long-term behavior} refers to user interest over longer time frames and includes, for example, preferences for particular shops or item genres.
For these two behavior types, we first present distinct strategies to discretize them into sequences of {\it behavioral words}. Compared to modeling single user actions sequentially, the proposed discretization leads to better generalization.
The token representation of these behavioral words is computed by concatenation and averaging of the word embeddings of the attribute IDs (\eg, shop, price, or product genre) of each action, and this is followed by the summation of token, position and segment embeddings.
These representation vectors are then aligned with the word embeddings of user categorical attributes as the input to UserBERT.
With this input, we design a novel \emph{pretext} task,  \emph{masked multi-label classification}.
The UserBERT model is pretrained via optimizing the multi-label classifications of the multiple attributes in the masked behavioral words.

Despite the parallels between user behavior and sentences, there are substantial differences and challenges in designing the learning procedure in a consistent way. 
Our model is able to deal with heterogeneous user behavior data, and achieve generalization via effective tokenization and the pretraining task.
The UserBERT model explores integrating various types of user data in a unified architecture and learning generic representations with self-supervised signals.
In our experiments, the pretrained model is fine-tuned on three different real-world tasks: user targeting, user attribute prediction, and next purchase genre prediction.
The results show that UserBERT outperforms task-specific modeling and multi-task learning based pretraining.

Our contributions are summarized as follows:
\begin{itemize}
    \item We propose UserBERT to pretrain user representations by making the analogy of actions in behavior sequences to words in sentences. We eliminates the need for collecting additional user annotation.
    \item UserBERT adopts a unified model architecture to enable simultaneous learning from  heterogeneous data, including long-term and short-term behavior as well as demographic data.
    \item We design the discretization of user raw data sequences, the generation of the input representation and a novel pretext task for pretraining.
    \item We evaluate UserBERT in extensive experiments. Compared with task-specific models without pretraining and multi-task learning based pretraining models, the proposed model achieves higher accuracy on three real-world applications.
\end{itemize}

\section{Related Work}

\subsection{Pretraining and Transfer Learning}

Recent studies have demonstrated that pretraining on large, auxiliary datasets followed by fine-tuning on target tasks is an effective approach~\citep{Oquab2014LearningAT, donahue14decaf,Hendrycks2019UsingPC,Ghadiyaram2019LargeScaleWP}.
Multi-task learning has been one of the commonly adopted approaches for pretraining due to its ability to improve generalization \citep{Zhang2017Survey,Ruder2017AnOO,gong2020profile}. It is shown that the pretrained MTL models can boost performance even when transferred to unseen tasks \citep{liu2015representation, Ni2018PerceiveInDepth}. 
Despite its success, MTL still has many challenges, such as negative transfer and the learning adjustment between different tasks \citep{Guo2018DynamicTP}. Also, MTL requires large amounts of well-annotated labels to produce satisfying outputs.
There are two common forms of adaptation when transferring the pretrained models to a given target task, \ie, \emph{feature-based} in which the pretrained weights are frozen, and directly \emph{fine-tuning} the pretrained model \citep{peters2019tuneornot}. 
We fine-tune pretrained models in our experiments.

\subsection{Self-Supervised Learning}
Deep learning models can already compete with humans on challenging tasks like semantic segmentation in the CV area \citep{he2015surpassing} as well as a few language understanding tasks \citep{liu2019mt-dnn}. 
However, such success relies on adequate amounts of quality training data, which can be expensive to obtain \citep{Kolesnikov2019RevisitingSV}. As a result, a lot of research efforts aim to reduce dependency on labeled data.  
Self-supervised learning (SSL), a subclass of unsupervised learning, has been drawing more attention since the recent advances in the NLP field.
Instead of using supervision signals, SSL only requires unlabeled data and trains models via formulating a \emph{pretext} learning task. 
There are two main types of pretext tasks: context-based \citep{Pathak2016ContextEF, norooziECCV16, sermanet2018tcvideo, Wu2019SelfSupervisedDL} and contrastive-based \citep{hjelm2019learning, chen2020simple}.
BERT \citep{Devlin2019BERTPO}, which our model is built upon, learns the contextual information through bi-directional transformers~\citep{Vaswani2017attenionisall} in a self-supervised manner.

\subsection{User Modeling}

To build user-oriented machine learning applications, a key challenge is finding an expressive representation of user bevhavior data, so that models can make accurate predictions.
For that reason, much effort has been spent on data preprocessing and transformations \citep{zhu2010clickmodel}.
Deep learning models have successfully mitigated the dependency on human efforts due to its ability to capture underlying representations in raw data \citep{zhou2018din, Li2020ASO}. However, these models need massive supervision signals for training, and they are mostly designed for specific tasks like recommendation \citep{pei2019reranking, sun2020bert4rec} and click-through rate prediction \citep{Zhou2019DeepIE}.

\begin{figure*}
  \centering
  \includegraphics[width=0.85\textwidth]{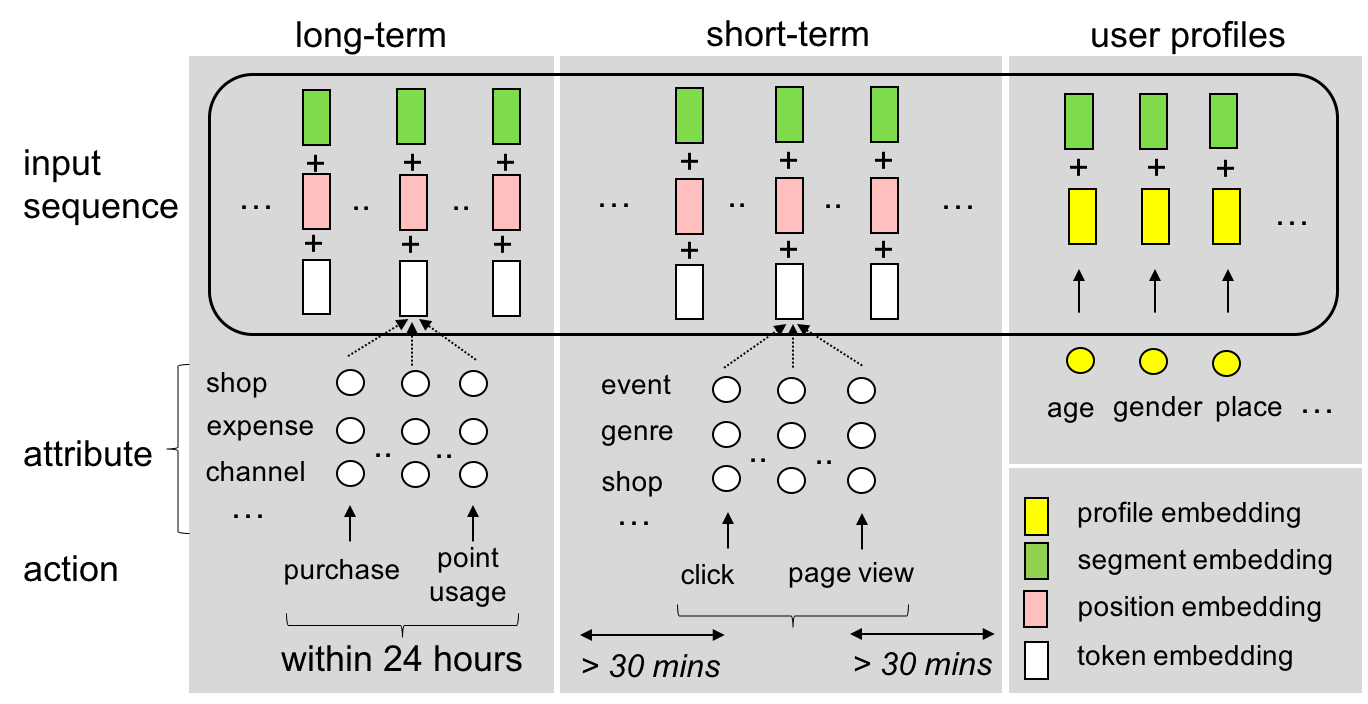}
  \caption{ \bf {Tokenization and input representation of long-term and short-term user behavior and attribute data.} \it To form behavioral words, we discretize long-term behavior into 24-hour intervals, and segment short-term sequences when there is a break of longer than 30 minutes between two actions. 
  The word embeddings of the attribute IDs of each action are first concatenated. Then, the token representation for one time interval is constructed by the mean of all action embeddings within the interval.
  The representation in the sequence is the sum of token embeddings and the embeddings for encoding position and segment.}
  \label{fig:tokenization}
\end{figure*}

Despite the success of these deep learning models, they fail to generate promising results for real-world industrial tasks with limited labeled data.
To deal with this issue, the methodology that pretrains universal user representations on massive user data, and then fine-tunes them for downstream tasks is explored. The goal is to learn a universal and effective representation for each user that can be transferred to new tasks \citep{Ni2018PerceiveInDepth, gong2020profile}.
However, MTL-based pretraining still requires the collection of user labels. Also, it is limited by inherent shortcomings to achieve optimal results \citep{Kendall2018MultitaskLU, Guo2018DynamicTP}. 

Recent work proposes learning user representations in a self-supervised way. For instance, PTUM applies Masked Behavior Prediction and Next K Behaviors Prediction to pretrain user models \citep{Wu2020PTUMPU}. CL4SRec uses a contrastive learning framework and proposes three data augmentation methods to construct contrastive tasks for pretraining \citep{Xie2020ContrastivePF}.
However, none of these works consider the intrinsic discrepancies of user behavior types. Also, the pretraining that sequentially models every single user actions is interfered with the randomness of user behavior, and fails to learn underlying user preferences.



\section{The Proposed Approach}
In this section, we first briefly review the BERT model, and then elaborate on how to extend it to user data including behavior sequences and demographic attributes. 

\subsection{The BERT Model}

BERT is a language representation model that pretrains deep bidirectional representations by jointly conditioning on both left and right contexts in all encoding layers \citep{Devlin2019BERTPO}. The input to the BERT model is a sequence of tokens that can represent both a single text sentence and a pair of sentences. These discrete tokens consist of words and a set of special tokens: separation tokens (SEP), classifier tokens (CLS) and tokens for masking values (MASK).
For a token in the sequence, its input representation is a sum of a word embedding, the embeddings for encoding position and segment. 

The BERT model is pretrained with two tasks, masked language modeling (MLM) and next sentence prediction. In MLM, the input tokens are randomly masked and the BERT model is trained to reconstruct these masked tokens. In detail, a linear layer is learned to map the final output features of the masked tokens to a distribution over the vocabulary and the model is trained with a cross entropy loss. In next sentence prediction, the inputs are two sampled sentences with a separator token SEP between them. The model learns to predict whether the second sentence is the successor of the first. A linear layer connecting the final output representations of the CLS token is trained to minimize a cross entropy loss on binary labels.
Many recent research works focus on extending the BERT model to areas beyond NLP, and successfully achieved state-of-the-art results \citep{Sun2019VideoBERTAJ,lu2019vilbert, Su2020VL-BERT, di2020imagebert}.

\begin{figure*}
  \centering
  \includegraphics[width=0.7\linewidth]{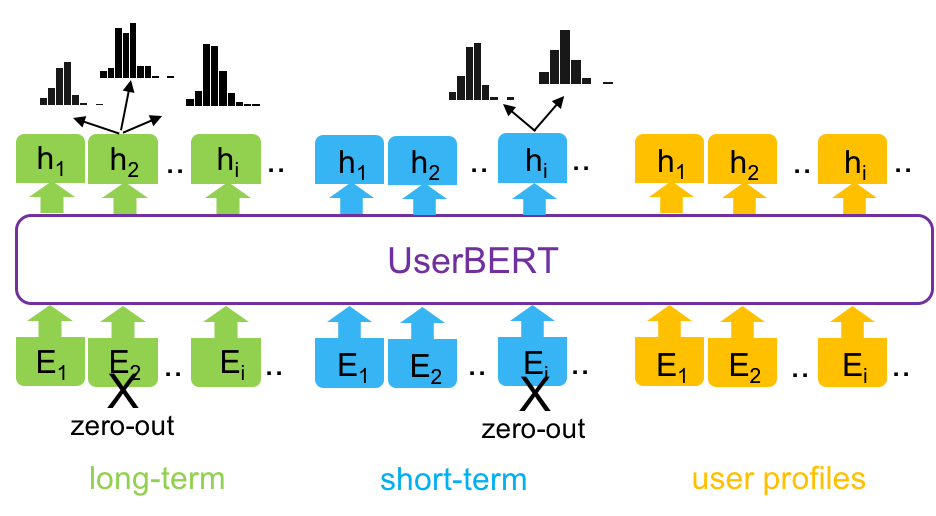}
  \caption{ \bf {Pretraining UserBERT.} \it The behavioral word representation vectors E\textsubscript{i}, in input sequences are randomly masked (zeroed-out), and the masked input is passed through UserBERT. 
  The model is trained to reconstruct the attributes in these masked words. 
  For each attribute, an output layer is connected to the final hidden representations h\textsubscript{i}, at the masked positions, and is learned by minimizing the multi-label classification loss.}
  \label{fig:pretraining_diagram}
\end{figure*}

\subsection{UserBERT}
\subsubsection{Tokenization of User Behavior Sequences}
Our goal is to learn generic user representations that characterize users based on their long-term preferences as well as recent interests.
We decide not to sequentially model single actions in long-term and short-term user data.
While such modeling is suitable for certain tasks, it is susceptible to overfitting when learning generic user representations.
Instead, we learn from a sequence of clustered user actions, in which a cluster represents a routine or a spontaneous interest.
Customers often make online purchases with specific intentions, \eg, shopping for a shirt, comic books, or a gift for Mother's Day.  
Many customers have long-standing preferences for particular stores and sales are heavily impacted by seasonality.
These continuous or related actions form a 'word' in a behavior sequence.
Similarly, we consider the same regarding short-term user behavior. Users commonly browse web content, moving between pages on an e-commerce site. During this time period, in order to capture the user's interest, we aim to estimate the theme or product genre rather than the specific order of individual actions.

Therefore, we first need to segment raw action data into a sequence of 'behavioral words' for each user, analogous to words in a sentence. 
As described by Figure \ref{fig:tokenization}, we adopt different approaches for long-term and short-term data.  Data representing long-standing user preferences is segmented into 24-hour intervals from 4 AM of one day to 4 AM of the next day. 
Short-term data is segmented if there is a time interval larger than 30 minutes between two actions, similar to the processing steps in \citep{grbovi2018rtembeds}.

\subsubsection{Input Representations}
In order to enable bidirectional representation learning, we transform the behavioral word sequence into a sequence of input embeddings. 
We first introduce the concept of \emph{action} and \emph{attribute} in user behavior data: The action indicates what a user does, \eg, making a purchase or obtaining points for using a service, while the action attributes include, for example, the type of the action, the shop name, the item genre, and price range, as shown in Figure \ref{fig:tokenization}. 
We choose different actions and their corresponding attributes in our dataset to represent long-term and short-term user behavior, and propose separate tokenization strategies for them since we expect to extract inherent user preferences from regular routines over longer time periods, and short-term interests from recent, temporary interactions. In combination with user attribute data, the learned representations are comprehensive and expressive.

To generate input representations, all attribute IDs are first mapped to fixed-length continuous vectors. 
These attribute vectors are concatenated for each action, obtaining an action embedding. Subsequently, the token representation for one time interval is obtained by taking the mean of all action embeddings within this interval. 
Finally, the input embedding vector is obtained by summing the token embeddings and the embeddings for encoding position and segment. 
Segment embeddings are used to differentiate the different types of user data, \ie, long-term user behavior, short-term behavior, and user profile data.
Long- and short-term user data share the same processing steps above, but each has their own definitions for token position.
For long-term sequences we use the number of days counted from the starting point of the collected training data, for short-term data it is the number of hours.
In order to incorporate non-temporal user attribute data to our modeling, we consider categorical attributes like gender as tokens in the user input sequence.
Continuous-valued attributes, such as age, are segmented by heuristics and converted into categorical attributes.
After mapping attributes to word embedding vectors, these are added to the segment embedding. Note that there is no position embedding for user-attribute embeddings since no order information needs to be captured for these user attributes.
The input sequence for each user is formed by aligning the generated representation vectors of user behavior as well as the embeddings of user attributes, see Figure \ref{fig:tokenization}. 


\subsubsection{Pretraining Tasks}
The generated input sequences allow us to make minimal changes to the BERT architecture and follow the practice in \citep{Devlin2019BERTPO}. 
We then pretrain our model to learn bidirectional representations.
While the language modeling task seems to naturally apply to our setting, reconstructing the masked 'behavioral words' requires modification since these words contain an assembly of user actions rather than individual words used in the original BERT model.
We implement \emph{masked multi-label classification} to predict multiple attributes in the masked behavioral words. More precisely, for each target attribute in a masked token, a linear layer is connected to the final representations which maps to a distribution over the attribute vocabulary, as illustrated in Figure \ref{fig:pretraining_diagram}. For one masked token, the training loss is the sum of cross entropy losses of all user attribute predictions, \eg, the prediction of shop IDs, genre IDs, etc. The final loss for one input sequence is the sum of the losses of all masked tokens. 

\begin{table*}[ht]
    \caption{Actions and attributes in user behavior data}
    \label{tab: experiment_data}
    \begin{center}
    \resizebox{0.8\linewidth}{!}{
        \begin{tabular}{l  l  l} 
        \toprule
        & \bf Actions  & \bf Attributes (with vocabulary size) \\ 
        \midrule 
        \multirow{2}{*}{long-term   $\quad$}      &\multirow{2}{*}{purchase, point usage} & action type (2), channel (742), expense range (17),\\
         & & shop (85,124), genre (11,438), hour (24)\\
        \midrule 
        \multirow{2}{*}{short-term}  & \multirow{2}{*}{click, search, page view} & action type (3), shop (40,804),\\
         & & genre (10,386), device type (2)\\
        \bottomrule
        \end{tabular}
        }
    \end{center}
\end{table*}

For masking input tokens, we follow a similar process as BERT: 15\% of tokens are selected uniformly, where 80\% of the time the token is zeroed-out and remains unchanged otherwise.
We distinguish between three segments of behavioral words from the three types of user data, \ie, long-term, short-term and user attributes. For long and short-term segments, we apply the masking-prediction for pretraining our model, while we do not mask user attributes.
To pretrain UserBERT, we first randomly sample a mini-batch of raw user sequences. Then, they are tokenized and transformed to input representations, which is followed by the masking step. In the end, the masked sequences are passed through the model, and the model is trained by minimizing the prediction error for reconstructing what attributes are inside the masked tokens. For each attribute type, a linear layer is learned to map the hidden representations of masked tokens to distributions over its vocabulary for conducting the multi-label classification.

Let $i$ be a randomly sampled index for masking, $w_{i}$ and  $w_{\backslash i}$ be the masked behavioral word and the input after masking to the UserBERT. 
Also, let $n$ be the number of target attributes for reconstruction prediction, and $\mathit{f}^{k}(w_{\backslash i}|\theta)$ be the final output vector after softmax layer for $k$-th attribute in the masked $w_i$. 
The loss of the UserBERT model is:
\begin{equation}
\label{eq:loss_func}
L(\theta)=\mathbb{E}_{w\sim D, i\sim \{1,..,t\}}\sum_{k=1}^{n} L_{C\!E}(y_{i}^{k}, \mathit{f}^{k}(w_{\backslash i}|\theta)),
\end{equation}
where $w$ is a uniformly sampled input representation sequence from the training dataset $D$, 
$t$ is the total number of behavioral words in the long-term and short-term data,
$y_{i}^{k}$ is the ground truth binary vector for the $k$-th attribute with its corresponding vocabulary size in the masked $w_i$ and $L_{C\!E}$ is the cross entropy loss for the multi-label classification. 
Note that long-term and short-term user behavior have different types and number of attributes in actions.
With the pretrained models, we leverage them for fine-tuning on downstream tasks.

\section{Experiments}

We experimentally verify whether the proposed UserBERT model is able to yield generic user representations, and evaluate the performance when applying it to different tasks via transfer learning.

\subsection{Experiment Settings}

\subsubsection{Datasets}

Datasets are collected from an online ecosystem of a variety of  services, including an e-commerce platform, a travel booking service, a golf booking service and others. Customers can access all  services via a unique customer ID, and their activities across the ecosystem are linked together. 

We consider two actions as long-term user behavior. The first one is the purchase action on the e-commerce platform, and the second one is a point usage action. Points are earned whenever purchases are made, or when certain services are used. Points can be spent on any service within the ecosystem.
The {\it channel} attribute represents from which service users obtain points or where they spend points. 
We collected purchase and point usage data over a three-month period.
For short-term behavior, we focus on recent  activities on the e-commerce website, \ie, browsing and search history. The relevant actions are clicks, page views, and searches, collected over a shorter time period of seven days. More detailed information on actions and attributes in the experimental data are shown in Table~\ref{tab: experiment_data}.

The user attribute data is registered customer information such as age and gender. The unique number of users in the dataset is 22.5M, the number of daily purchase and point usage samples is approximately 5M, and the number of short-term data samples is approximately 50M. The data is preprocessed to generate user action sequences. 




\subsubsection{Target Tasks}
Our pretrained user model is trained in a general manner and can be adapted to a variety of user understanding tasks. We fine-tune the self-supervised pretrained model to three real-world downstream tasks that aim to improve customer experience. The datasets of the three target tasks are split 80-20 to create training and testing datasets for fine-tuning.

\begin{itemize}
    \item The \textbf{user targeting} task is to identify potential new customers for certain services or products, and it is formulated as a binary classification problem, indicating interest or no interest in a particular service. 
    Users who responded positively to a target service/product, \eg, directly via a purchase or indirectly by clicking on a banner, form the set of positive labeled data, while negative ones are uniformly sampled from the remaining set of users with a 3:1 ratio. A new dataset is collected after the time period of the data used for pretraining.
    \item The \textbf{user attribute prediction} task is predicting different user attributes, \eg, whether or not a customer owns a pet. It is also posed as a classification problem, where ground truth labels are obtained through questionnaires.
    \item The \textbf{next purchase genre prediction} task is a multi-class prediction problem with the aim to predict the next genre of items that a user will purchase  on the e-commerce platform. The dataset is created from the one-month user history following the pretraining time period.
\end{itemize}

\begin{figure*}[h]
    \centering
    
    \hspace*{-1.05em}%
    \begin{subfigure}[b]{0.27\linewidth}
        \includegraphics[width=\linewidth]{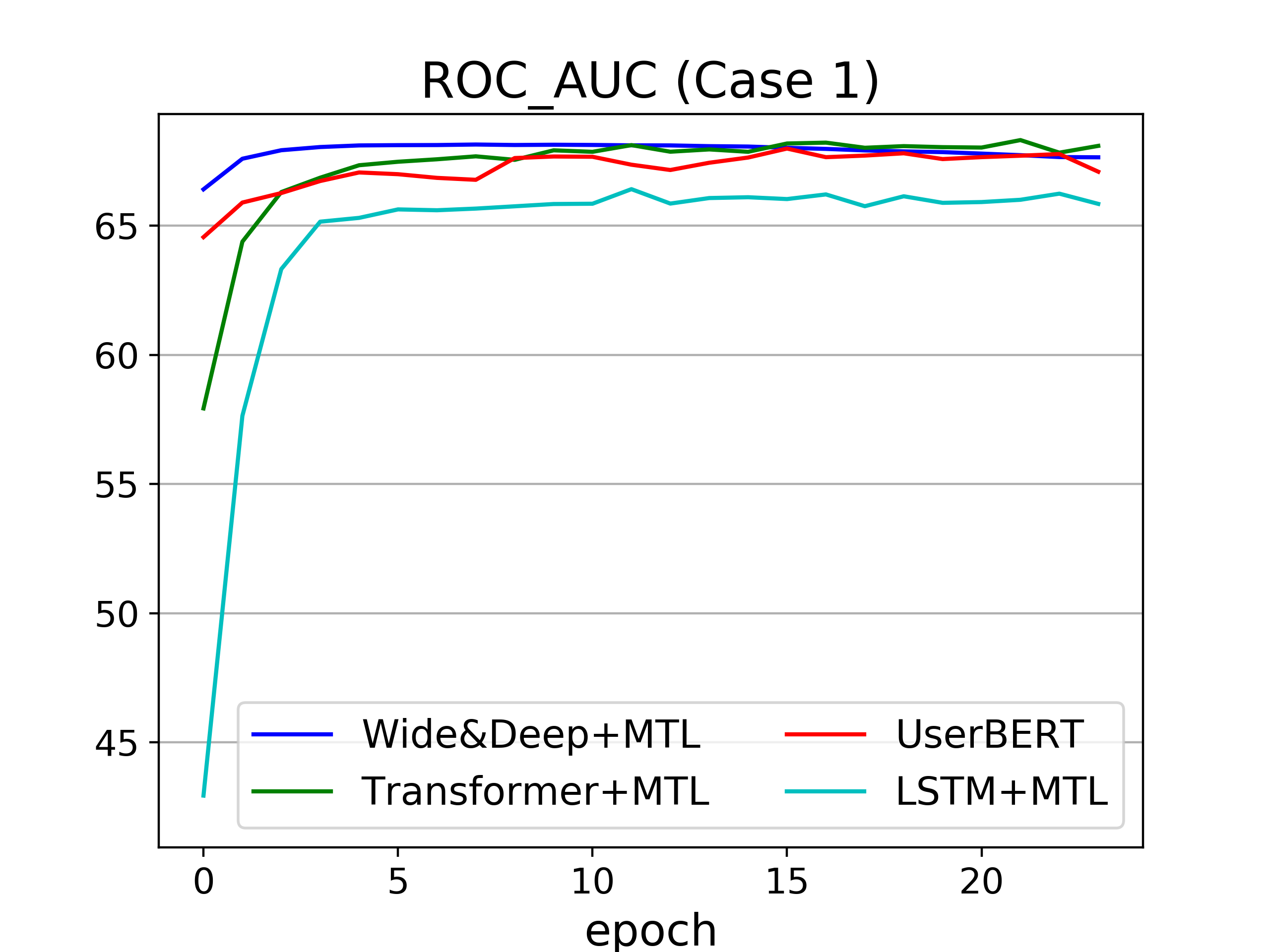}
    \end{subfigure}\hspace*{-1.05em}%
    \begin{subfigure}[b]{0.27\linewidth}
        \includegraphics[width=\linewidth]{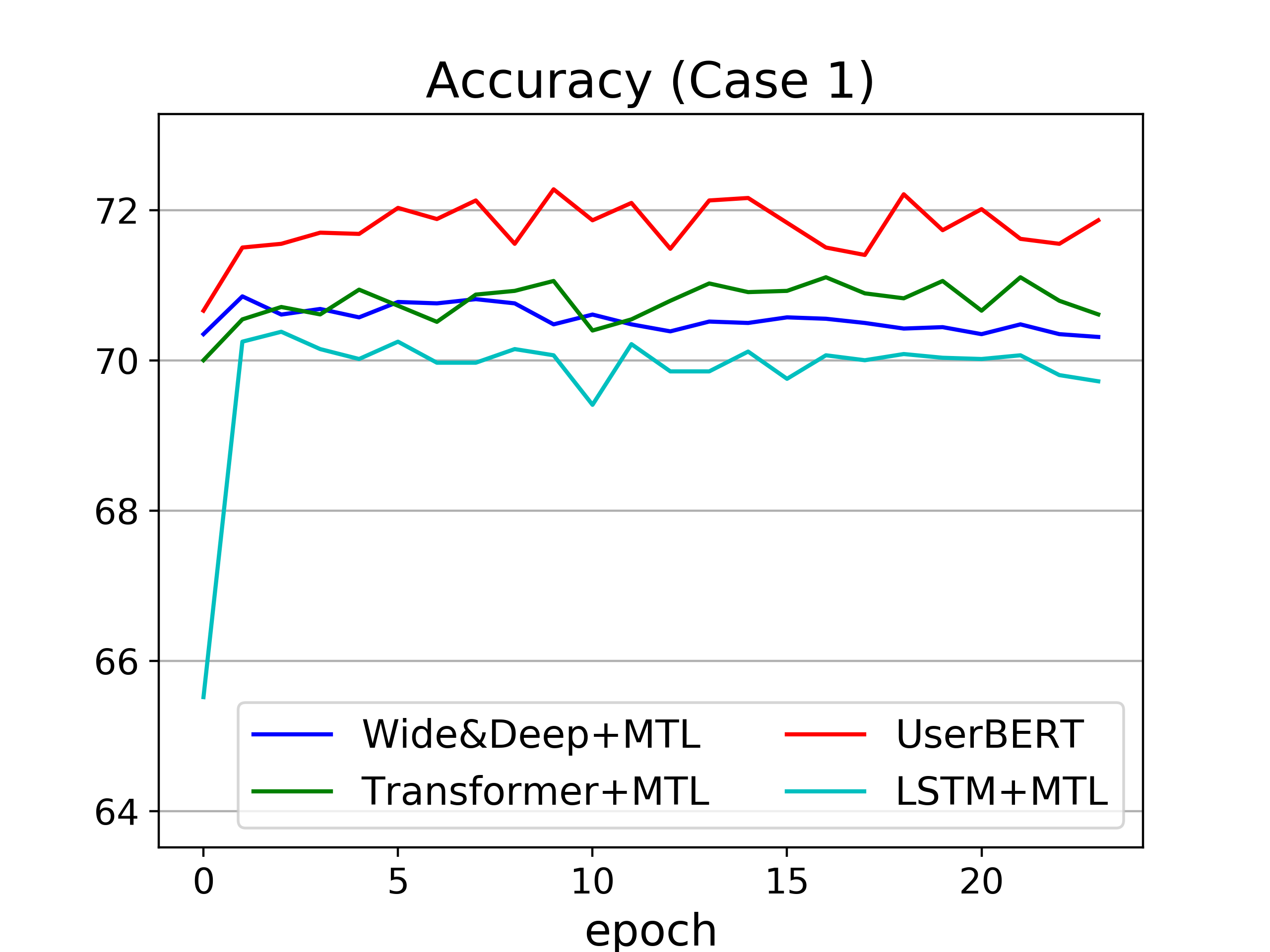}
    \end{subfigure}\hspace*{-1.05em}%
    \begin{subfigure}[b]{0.27\linewidth}
        \includegraphics[width=\linewidth]{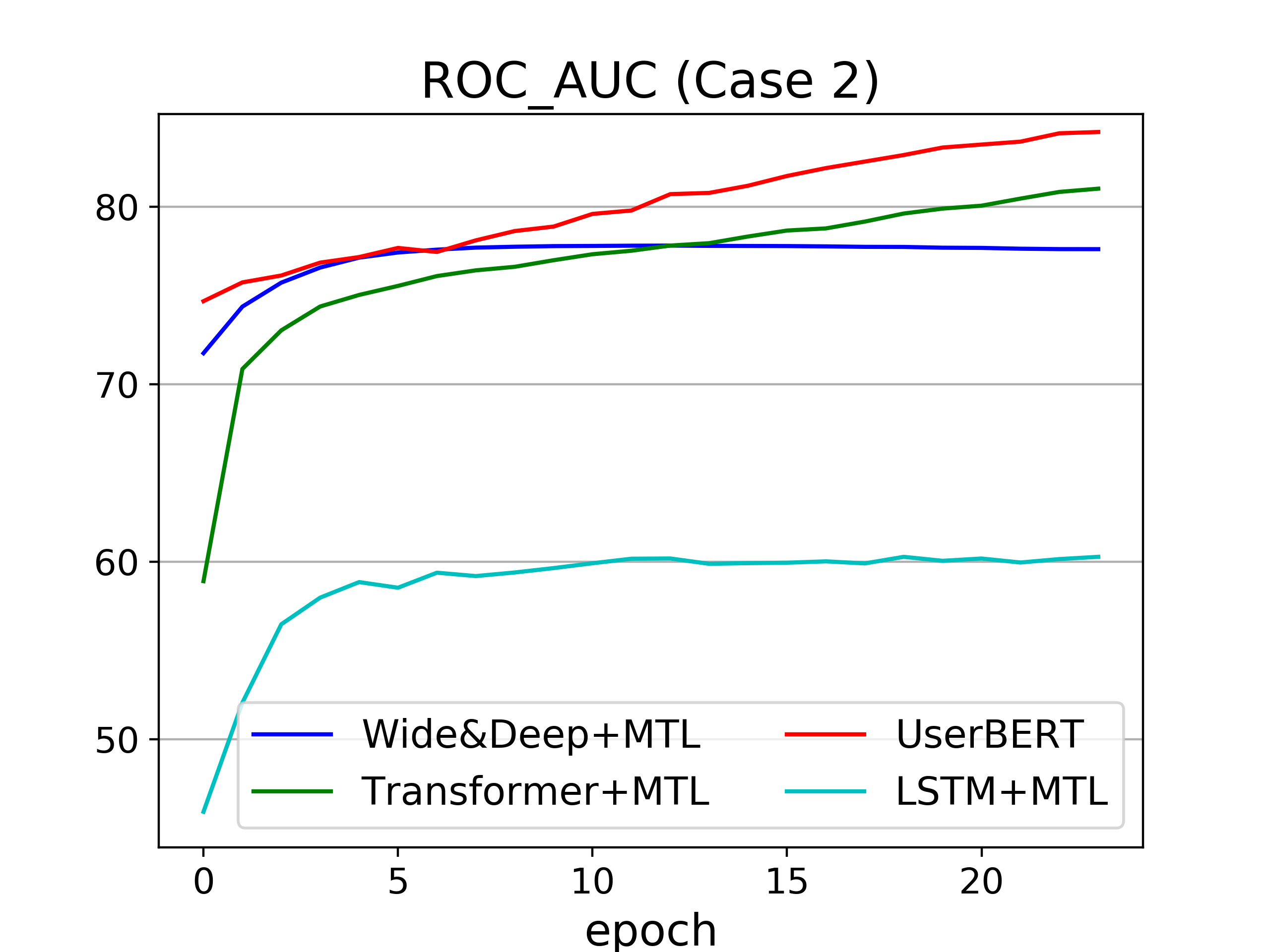}
    \end{subfigure}\hspace*{-1.05em}%
    \begin{subfigure}[b]{0.27\linewidth}
        \includegraphics[width=\linewidth]{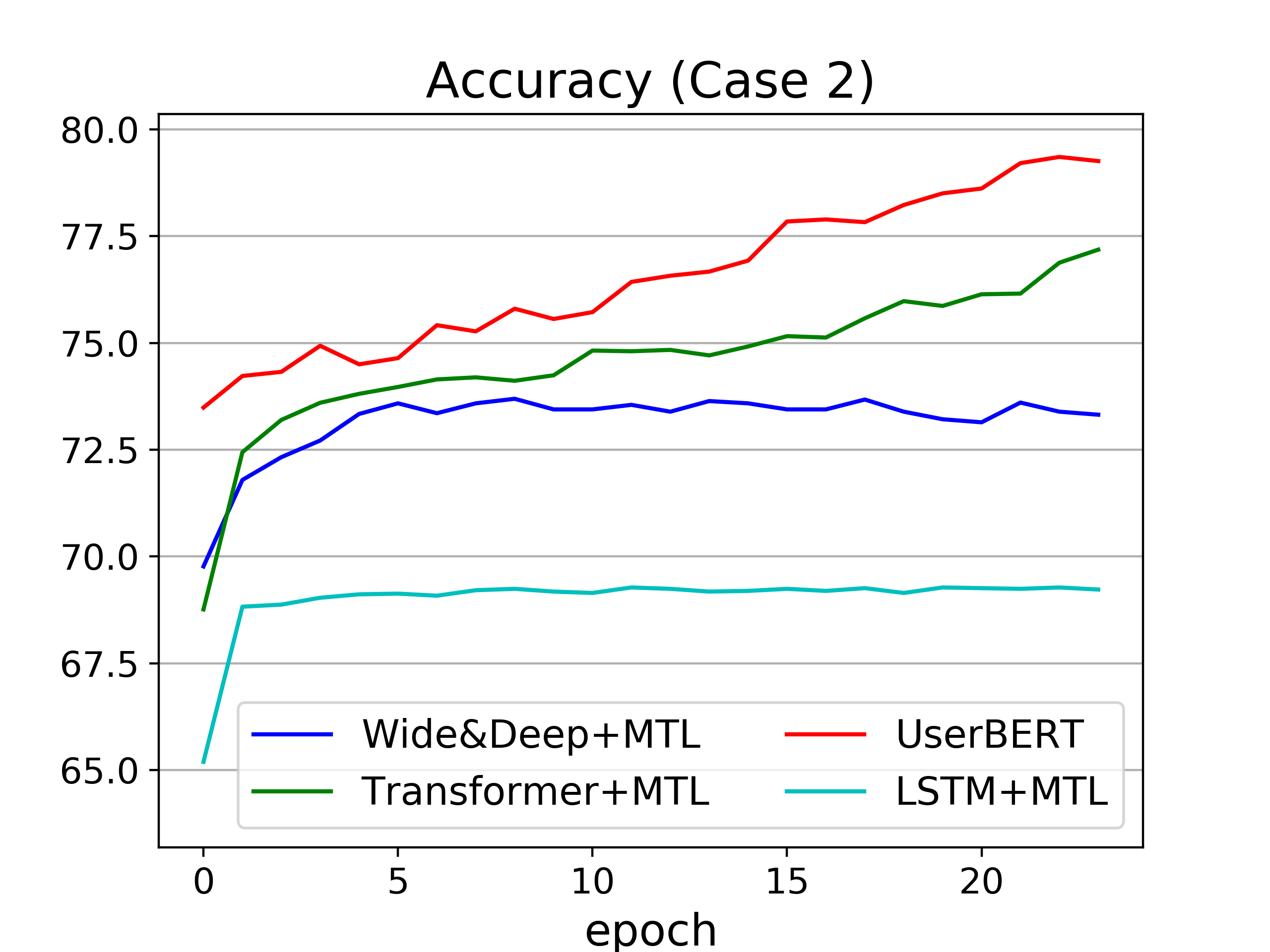}
    \end{subfigure}
    
    \caption{{\bf Model fine-tuning performance comparison for the {\em user targeting} task.} \it The charts show the results for task of predicting whether customers will user either of two different services (case 1 and case 2).
    For each case we plot the ROC AUC and accuracy metrics vs. the number of training epochs.}
    \label{fig:user_targeting_comparison}
\end{figure*}

\subsubsection{Model Baselines}
UserBERT is compared to direct modeling without pretraining and to multi-task learning (MTL)-based pretraining.
The MTL models apply a multi-tower architecture in which each tower encodes one type of user data.
For the MTL-based baselines, different types of user data are passed through corresponding encoders, and the encoded representations are combined at the last layer before connecting to multiple training tasks.
The dimension of the combined representations is set to 128 for all MTL models.

We collect user labels across the services in the ecosystem and train MTL models with 12 user attributes as pretraining tasks.  
These pretraining tasks classify the categories of user activities such as the usage frequency of certain services or attributes like type of occupation. 
We will not reveal the details of the tasks due to data governance.
By learning and sharing across multiple tasks, the yielded user representations are considered to be generalized and applicable for transferring to downstream tasks.

\begin{itemize}
    \item \textbf{Wide\&Deep+MTL}: 
The Wide\&Deep model is selected for comparison because it represents a traditionally applied approach for modeling e-commerce data. Although the model cannot directly handle user behavior data as sequences, we generate fixed-length (1130-d) embeddings by aggregating behavior data and inputting them to the deep part of the model~\citep{cheng2016widedeep}. Categorical user attribute data is mapped to word embeddings and concatenated before feeding it into the wide part of the model. The wide part is a linear model, while the dimensions of the 4 hidden layers for the deep side are 512, 256, 256 and 128, respectively. 

    \item \textbf{LSTM+MTL}: 
LSTM networks are commonly used to model sequential data \citep{Ni2018PerceiveInDepth}. The same discretization and input generation are applied to long-term and short-term user behavior for this model. It is a 3-tower model,
in which two LSTM networks model the two types of user behavior and user attributes are modeled in the same way as the Wide\&Deep model.
The dimension of the hidden state in all LSTM encoders and the length limitation of both long-term and short-term data are set to 128.

    \item \textbf{Transformer+MTL}: 
The architecture is the same as the  LSTM+MTL model above but with two different Transformer encoders \citep{Vaswani2017attenionisall} to model long and short-term user data separately.
The length of input user behavior sequence to the encoders is limited to 128 as well. We pretrain the model via minimizing the summed cross entropy loss of the multiple training tasks.

    \item \textbf{UserBERT}: 
The proposed self-supervised learning based pretraining model, which
simultaneously learns from long- and short-term actions and user attributes. Pretraining is done by reconstructing attributes in masked tokens via multi-label classifications.
\end{itemize}

\subsubsection{Experimental Setup}

For UserBERT, we use the same notations as BERT, and set the number of Transformer blocks $L$ to 4, the hidden size $H$ to 128, and the number of self-attention heads $A$ to 4. The input sequence length of both long-term and short-term data is limited to 128.
For fair comparison, we pretrain all models using the Adam optimizer with a learning rate of $10^{-4}$ and batch size of 16. We fine-tune models using the same learning rate and a batch size of 128. Pretraining of 400K batches of the UserBERT model takes approximately 12 hours using our PyTorch implementation, running on two GeForce RTX 2080 Ti GPUs.

For fine-tuning each target task, the combined encoder representations of the MTL-based models are fed to an output layer, while the fine-tuning of UserBERT is done by connecting the hidden representations of the first token to an output layer for each task.
After plugging in task-specific inputs and outputs, we fine-tune the parameters of pretrained models end-to-end.


\subsection{Results}

\subsubsection{User Targeting}
\begin{table}[t]
    \caption{{\bf User targeting task.}  Best ROC AUC and Accuracy comparisons after fine-tuning.}
    \label{tab: user_targetting_comparison}
	\centering
	\ra{1.2}\resizebox{\linewidth}{!}{
    \begin{tabular}{l  c  c  c  c }
        \toprule
        \bf  & \multicolumn{2}{c}{\bf{ROC AUC}} & \multicolumn{2}{c}{\bf Accuracy} \\
        \midrule
        \bf Model & \bf{Case 1} & \bf{Case 2} & \bf{Case 1} & \bf{Case 2} \\ 
        \midrule 
        Wide\&Deep+MTL & 68.14 & 77.81 & 70.61 & 73.67 \\
        LSTM+MTL & 66.42 & 60.28 & 70.38 & 69.27\\
        Transformer+MTL & \bf68.31 & 81.21 & 71.11 & 77.19\\
        UserBERT & 67.98 & \bf 84.20 & \bf 72.28 & \bf 79.36\\
        \bottomrule
    \end{tabular}}
\end{table}

We show the results for two different services. The sizes of the datasets are 30,204 and 31,106 samples, respectively. Compared to the size of the pretraining dataset, the use cases of this task only have few labeled samples. Classification performance per epoch in terms of accuracy and ROC AUC are shown in Figure \ref{fig:user_targeting_comparison}. Table \ref{tab: user_targetting_comparison} compares the best ROC AUC and accuracy results for the same two cases. The LSTM model, which sequentially models user behavior, has relatively low accuracy, indicating that the sequential order of user actions does not provide useful information for this task.

From our experience, the user targeting task focuses on patterns from relatively static user preferences. The Wide\&Deep model shows competitive performance, achieving the highest ROC AUC for case 1, which is reasonable since our exploratory analysis indicates that user attributes are important features. 
The performance of the Transformer-based models reveal that the underlying explanatory factors for this task can be captured by attention networks. UserBERT outperforms other models in terms of accuracy by a substantial margin.

\subsubsection{User Attribute Prediction}

In general, it is challenging to predict user attributes because predictive signals in the behavior data are sparse. 
In other words, the target user attributes may not be strongly correlated to behavior data. Therefore, this prediction task evaluates the model's ability to discover hidden explanatory factors in the raw data.
We show experimental results for two use cases: one is to predict whether a user has a car, while the other one is to predict if a user is a parent. These two tasks are denoted as {\it has\_car} and {\it is\_parent}. 

\begin{figure}
    \centering
    \hspace*{-0.4em}%
    \begin{subfigure}[b]{0.55\linewidth}
        \includegraphics[width=\linewidth]{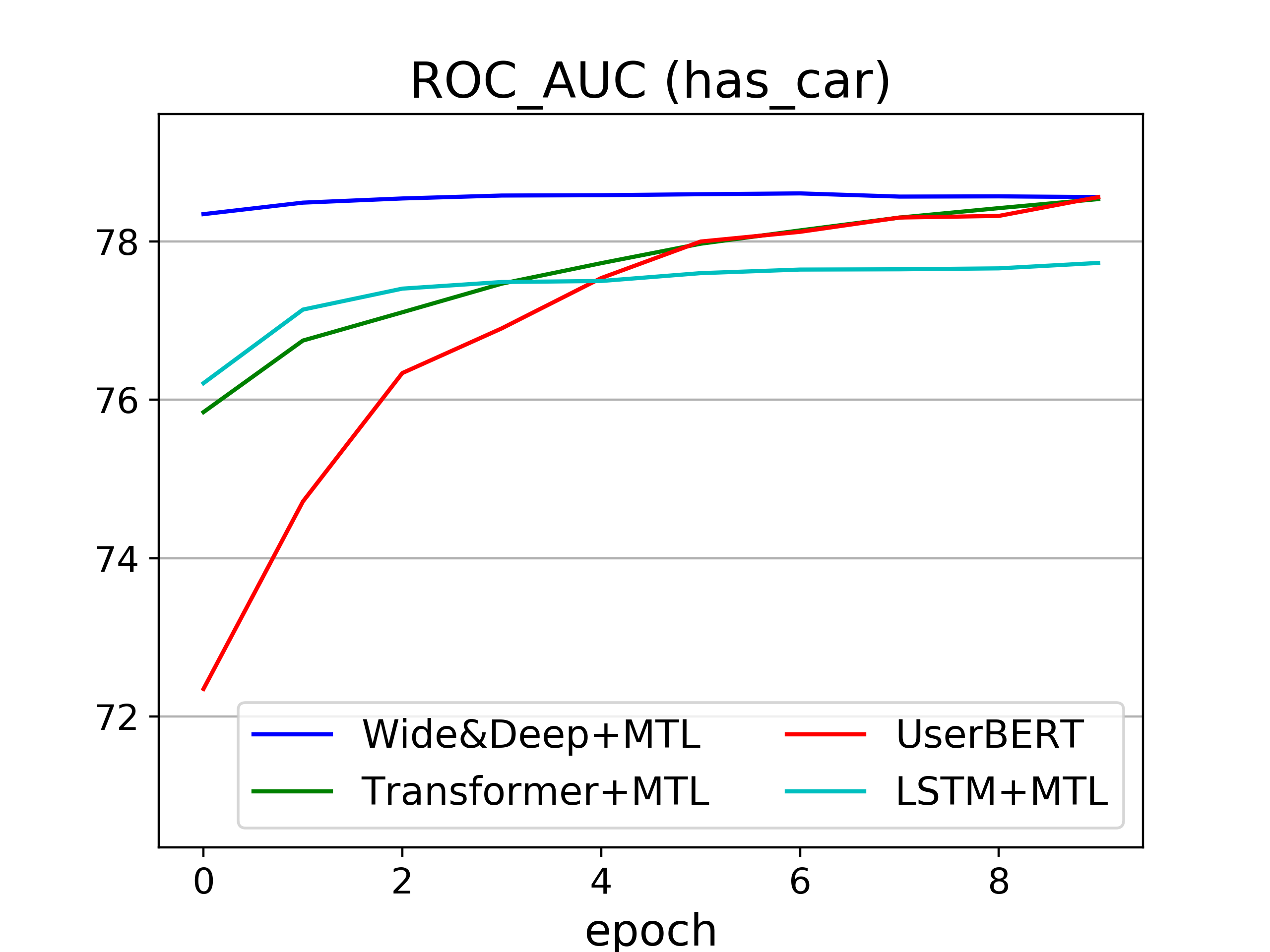}
    \end{subfigure}\hspace*{-0.8em}%
    \begin{subfigure}[b]{0.55\linewidth}
        \includegraphics[width=\linewidth]{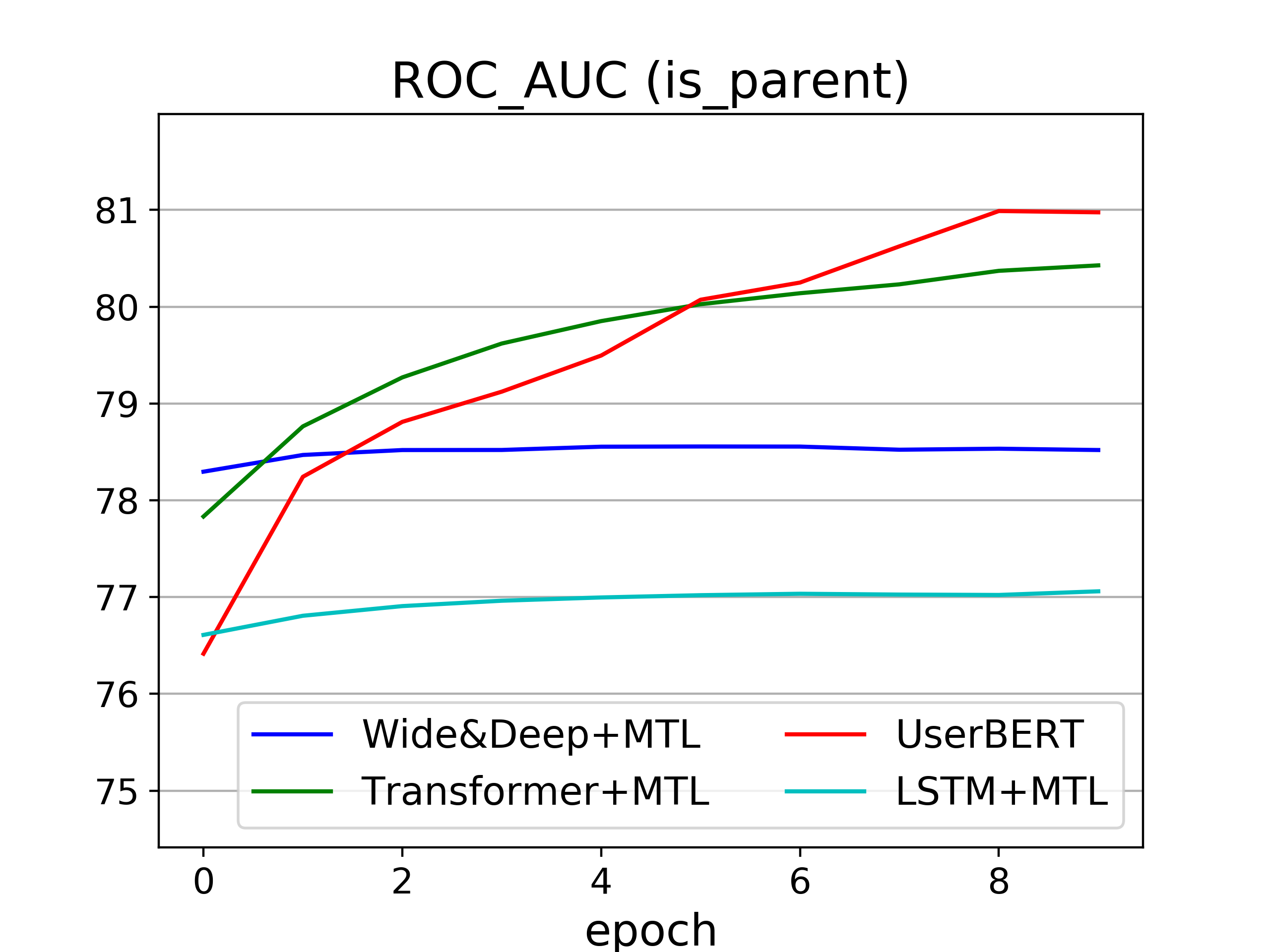}
    \end{subfigure}
    \caption{{\bf Model fine-tuning comparison on \emph{user attribute prediction} tasks.} \it  ROC AUC metric vs.  number of training epochs for different models on two different user attribute prediction tasks.}
    \label{fig:attribute_comparison}
\end{figure}

The dataset for the {\it has\_car} task contains 448,501 samples and the one for the {\it is\_parent} task contains 400,268. The classification results of 10-epoch fine-tuning are shown in Figure~\ref{fig:attribute_comparison}. Table~\ref{tab: attr_prediction_comparison} compares the best ROC AUC results in 10 epochs.
From the {\it has\_car} results, we observe that the Wide\&Deep model shows good initial performance, and during training other models eventually reach similar accuracy. We believe this is due to the fact that user features such as age and location are important features for this task. It seems challenging for models to extract other discriminative patterns from either long-term or short-term user behavior. On the other hand, whether a user is a parent or not seems to present different characteristics in terms of how they behave on an e-commerce or travel booking platform. These patterns can be captured by deep learning models like UserBERT and Transformer-based models. 
UserBERT is able to match and eventually outperform the baseline models.

\begin{table}[t]
    \caption{{\bf User attribute prediction.} Best ROC AUC comparison after fine-tuning.}
    \label{tab: attr_prediction_comparison}
	\centering
    \begin{tabular}{l  c  c}
        \toprule
        \bf Model & \bf{Has car} & \bf{Is parent} \\ 
        \midrule 
        Wide\&Deep+MTL & \bf 78.61 & 78.52 \\
        LSTM+MTL & 77.73 & 77.06 \\
        Transformer+MTL & 78.54 & 80.43 \\
        UserBERT & 78.56 & \bf 80.99 \\
        \bottomrule
    \end{tabular}
\end{table}

\subsubsection{Next Purchase Genre Prediction}

\begin{table}[t]
    \caption{{\bf Next purchase genre prediction. }Best mAP@10 comparison after fine-tuning.}
    \label{tab: next_genre_map_comparison}
	\centering
    \begin{tabular}{l  c}
        \toprule
        \bf Model & \bf mAP@10(\%) \\ 
        \midrule 
        Popular Genres & 4.22 \\
        Wide\&Deep+MTL & 7.65 \\
        LSTM+MTL & 8.13 \\
        Transformer+MTL & 8.62 \\
        UserBERT & \bf10.90 \\
        \bottomrule
    \end{tabular}
\end{table}

The dataset contains data from 586,130 users, and we fine-tune each pretrained model for 10 epochs. 
The mean average precision for the top 10 purchased genres (mAP@10) comparison is shown in Table \ref{tab: next_genre_map_comparison}. The UserBERT model outperforms baseline models by a large margin. This task requires understanding of both long-term preferences as well as recent interests of customers. Prediction models should be able to identify candidate genres from user habits over a longer time range, and then identify likely ones as prediction results from recent interest trends. More specifically, a model should understand how users typically use services in the ecosystem as well as what they are currently interested in.
The architecture of the baseline models learns from different types of user data separately and combines the last-layer representations for training. It fails to sufficiently capture the correlations.
In contrast, UserBERT benefits from the unified structure of the user data and captures more accurate correlations, not only within certain types of user behavior, but also between different behavior types via attention networks.

Since it is common that users make purchases from only a subset of genres, we also devised an intuitive but strong baseline that predicts the most popular genres, ranked by the total number of purchases, and compared it against all pretrained models. With an mAP@10 of 4.22\%, this model's accuracy is significantly lower, demonstrating the effectiveness of the pretrained models.

\subsection{Ablation Studies}

We perform additional experiments to better understand the effects of certain aspects of the pretraining and fine-tuning framework. More specifically, we analyze the effect of the pretraining step of UserBERT and how the number of labeled samples affects the performance of fully supervised pretraining methods.

\subsubsection{Effect of Pretraining}

\begin{figure}
    \centering
    \hspace*{-1.0em}%
    \begin{subfigure}[b]{0.56\linewidth}
        \includegraphics[width=\linewidth]{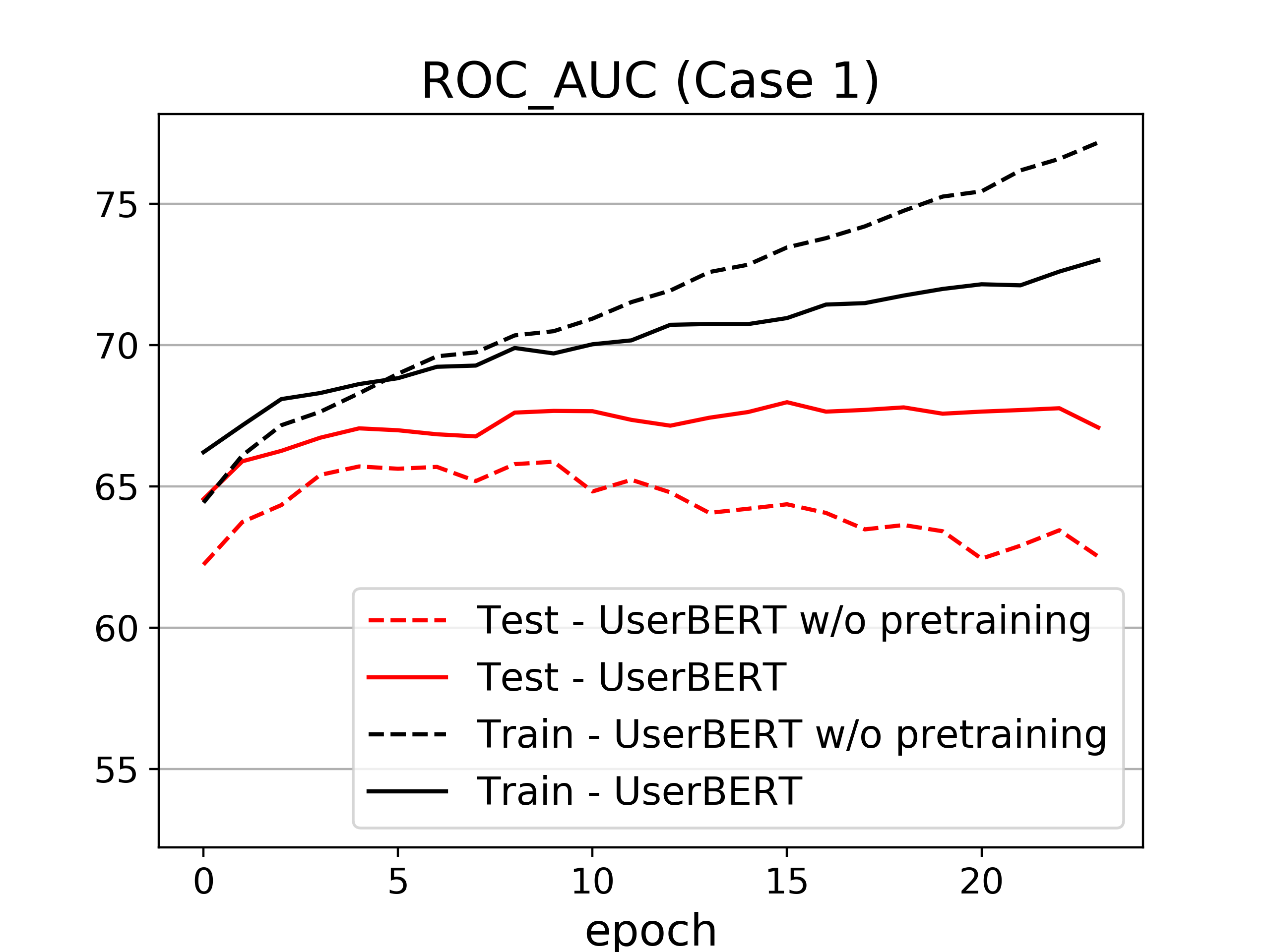}
    \end{subfigure}\hspace*{-0.8em}%
    \begin{subfigure}[b]{0.56\linewidth}
        \includegraphics[width=\linewidth]{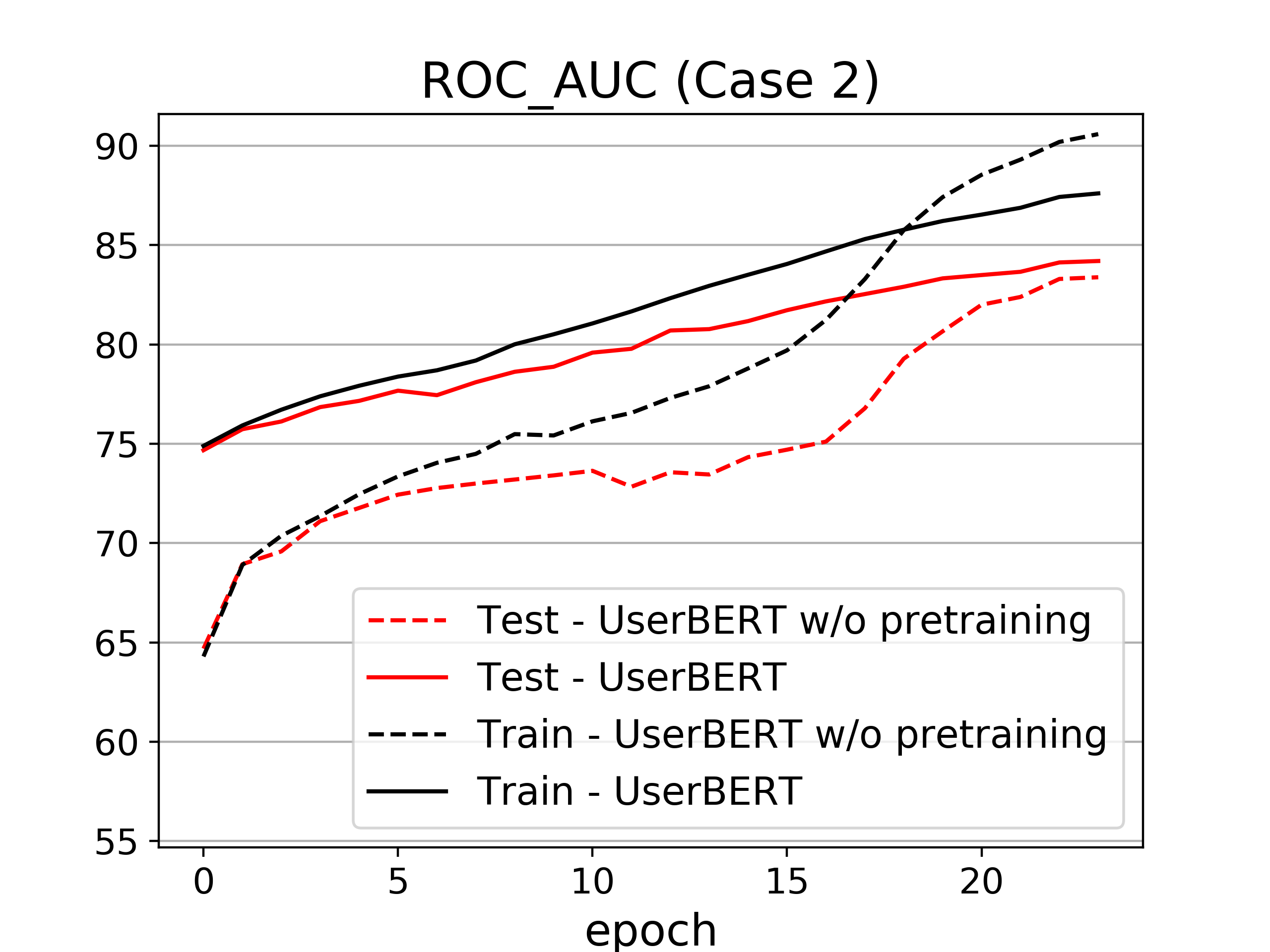}
    \end{subfigure}
    \caption{{\bf Effect of pretraining on UserBERT.} \it ROC AUC comparison of UserBERT with and without pretraining on the \emph{user targeting} task.}
    \label{fig:user_targeting_pretrainingVSnoPretraining}
\end{figure}

We directly apply UserBERT to the user targeting task without pretraining to verify whether it benefits from the pretraining step. The ROC AUC comparison between
UserBERT with and without pretraining is shown in Figure \ref{fig:user_targeting_pretrainingVSnoPretraining}. The pretrained models outperform the models trained from scratch significantly. 
This indicates that the pretraining step  extracts useful information that allows fine-tuning to boost performance for downstream tasks. 
From the error curves during training, we also observe that models tend to overfit quickly without pretraining. The pretrained UserBERT model achieves more generic user representations and yields significant accuracy improvements when adapted to new tasks.  

\begin{figure}
    \centering
    \hspace*{-1.0em}%
    \begin{subfigure}[b]{0.56\linewidth}
        \includegraphics[width=\linewidth]{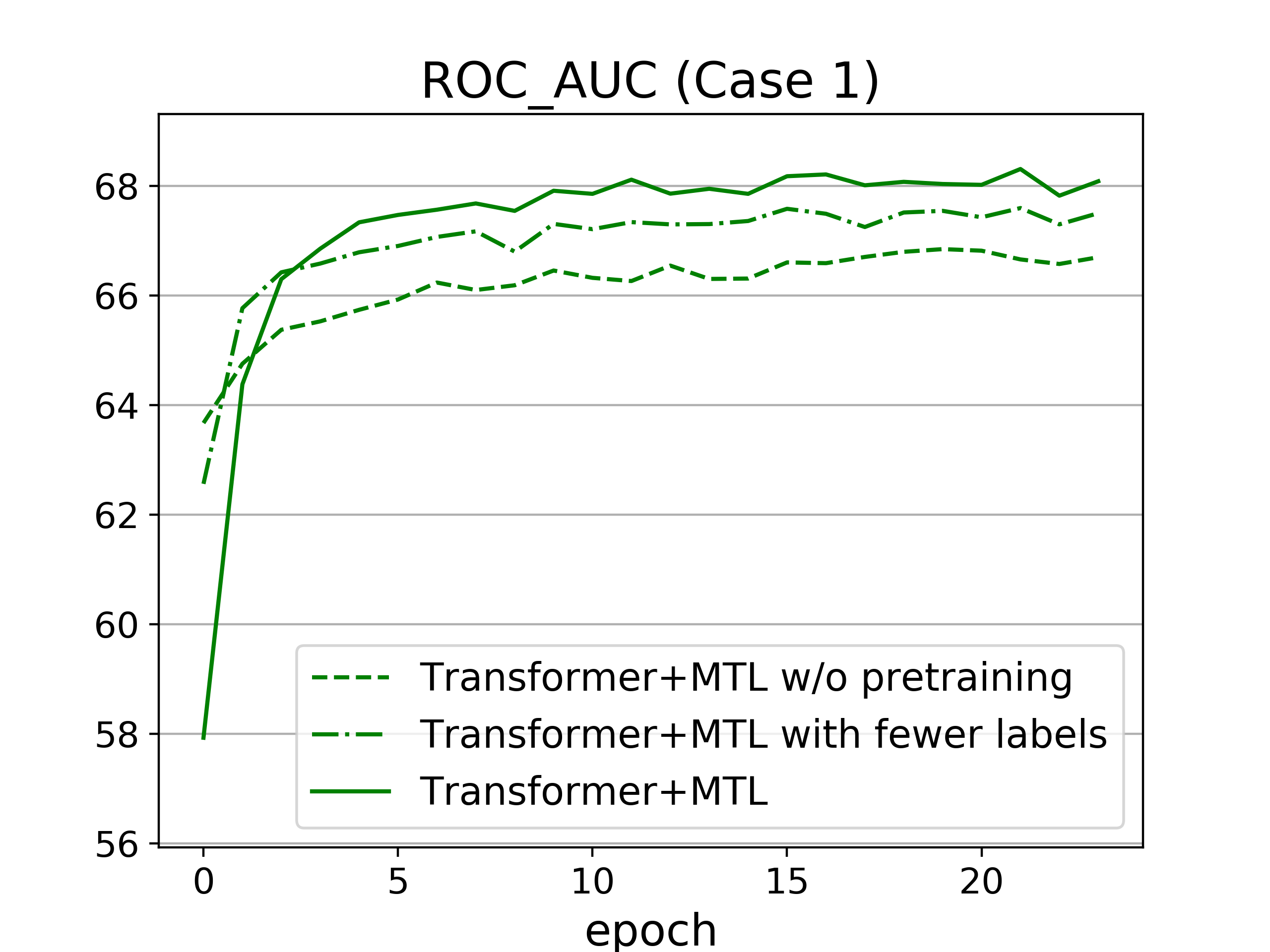}
    \end{subfigure}\hspace*{-0.8em}%
    \begin{subfigure}[b]{0.56\linewidth}
        \includegraphics[width=\linewidth]{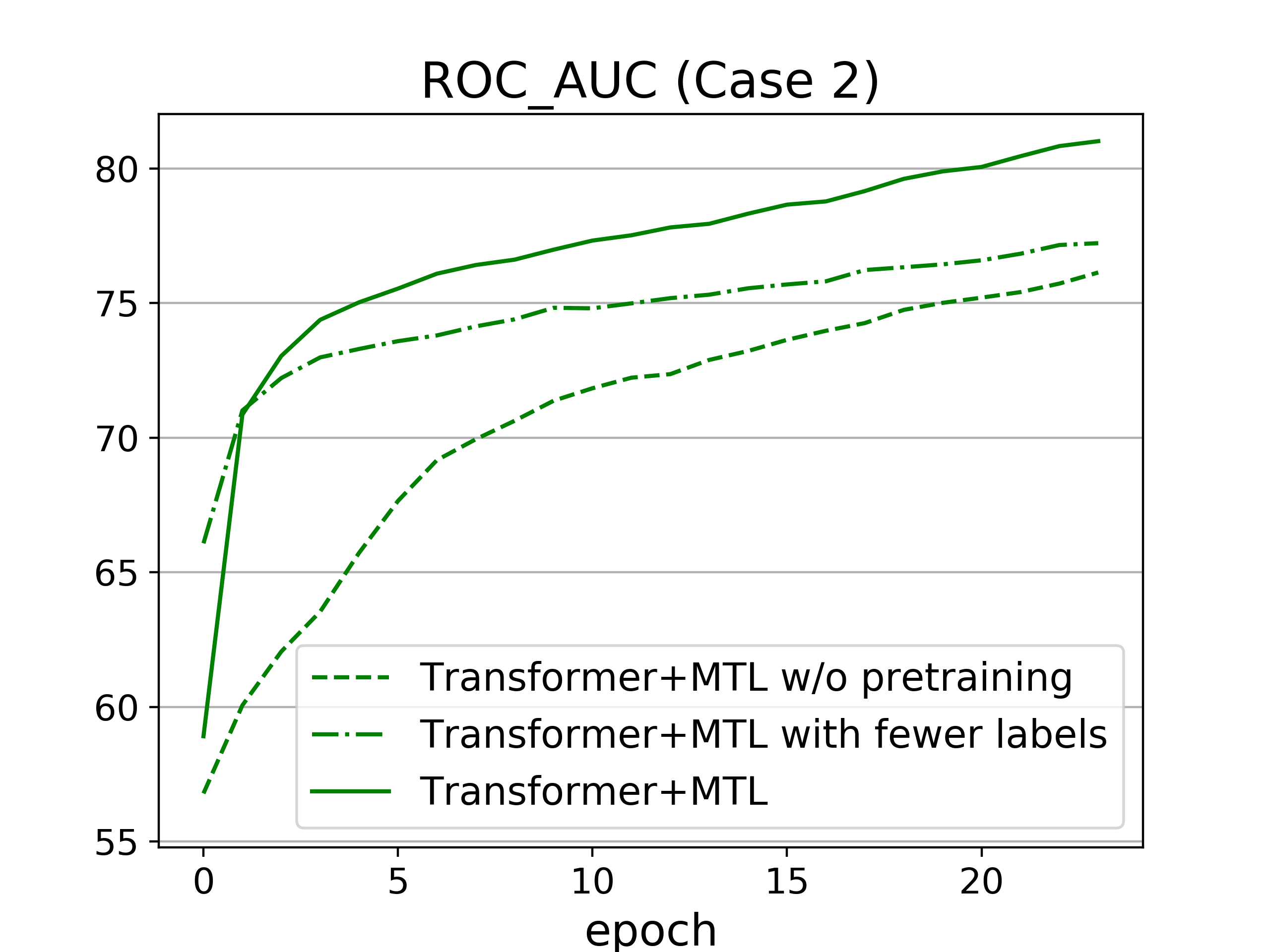}
    \end{subfigure}
    \caption{{\bf Effect of pretraining on Transformer+MTL.} Comparison between Transformer-based MTL models with no pretraining, pretraining with 30\% of labels, and pretraining with all labels on the \emph{user targeting} tasks.}
    \label{fig:user_targeting_fewer_labels}
\end{figure}

\begin{figure}[h]
    \centering
    \hspace*{-1.05em}%
    \begin{subfigure}[b]{0.55\linewidth}
        \includegraphics[width=\linewidth]{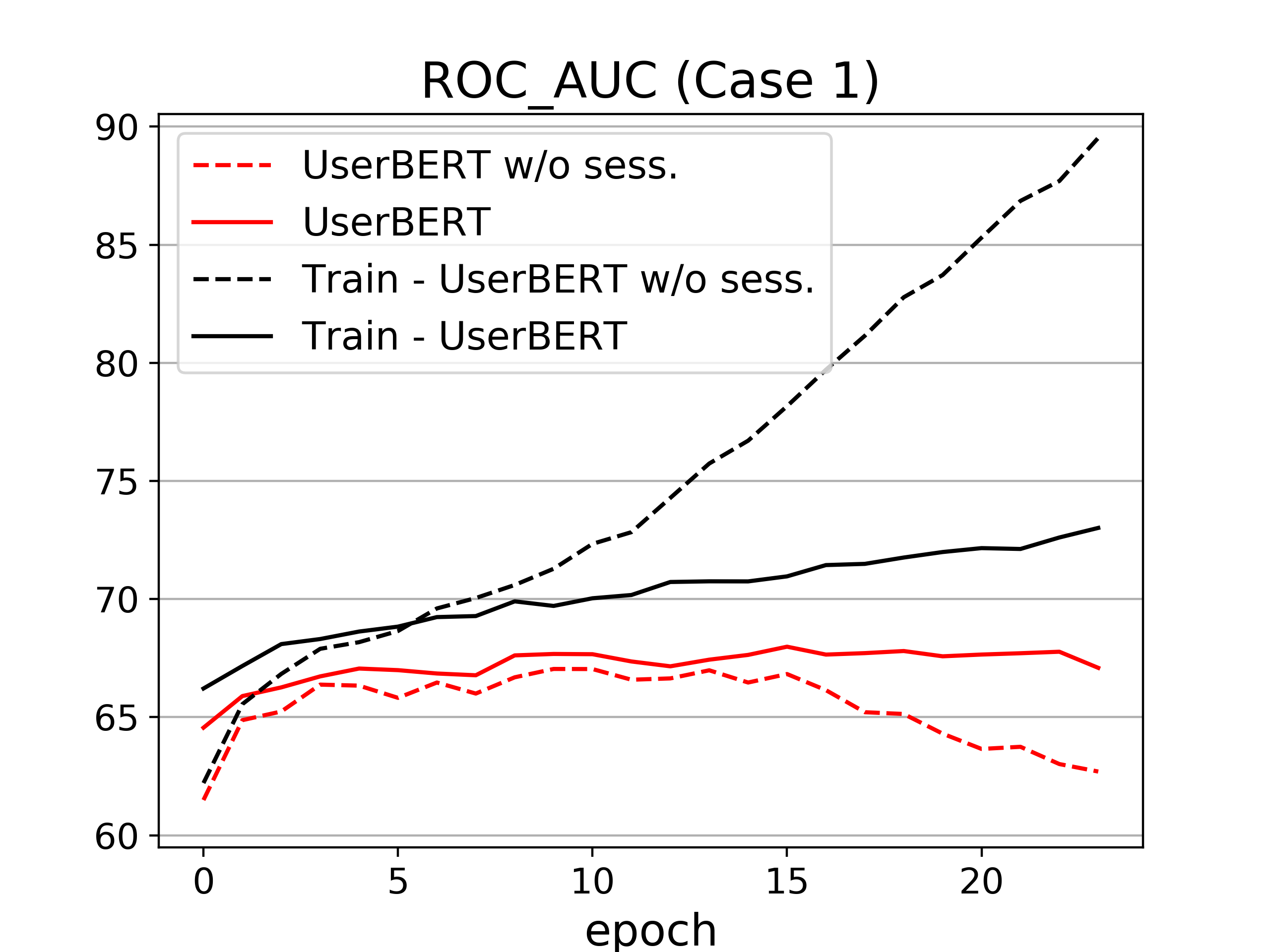}
    \end{subfigure}\hspace*{-1.05em}%
    \begin{subfigure}[b]{0.55\linewidth}
        \includegraphics[width=\linewidth]{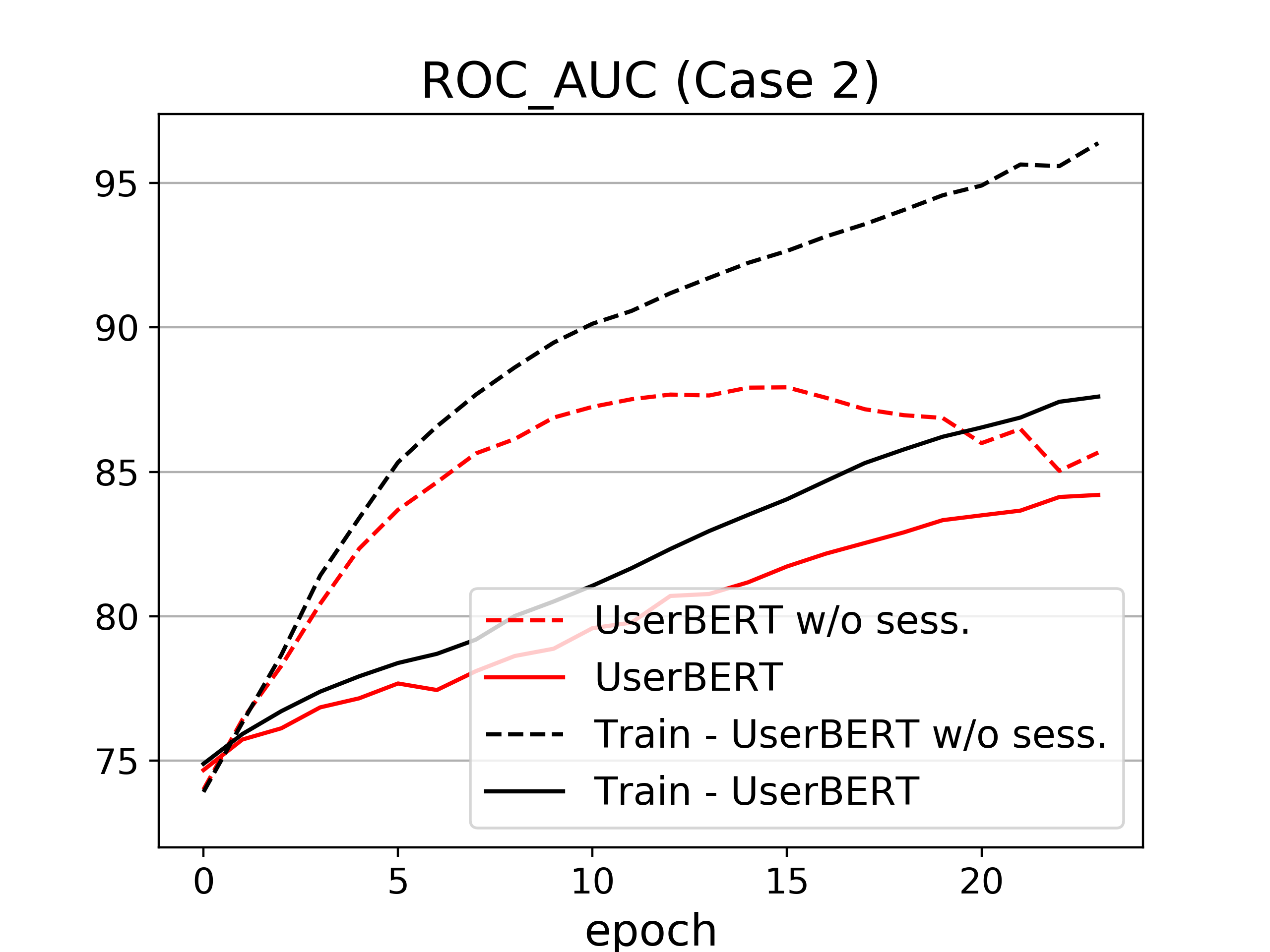}
    \end{subfigure}
    
    \hspace*{-1.05em}%
    \begin{subfigure}[b]{0.55\linewidth}
        \includegraphics[width=\linewidth]{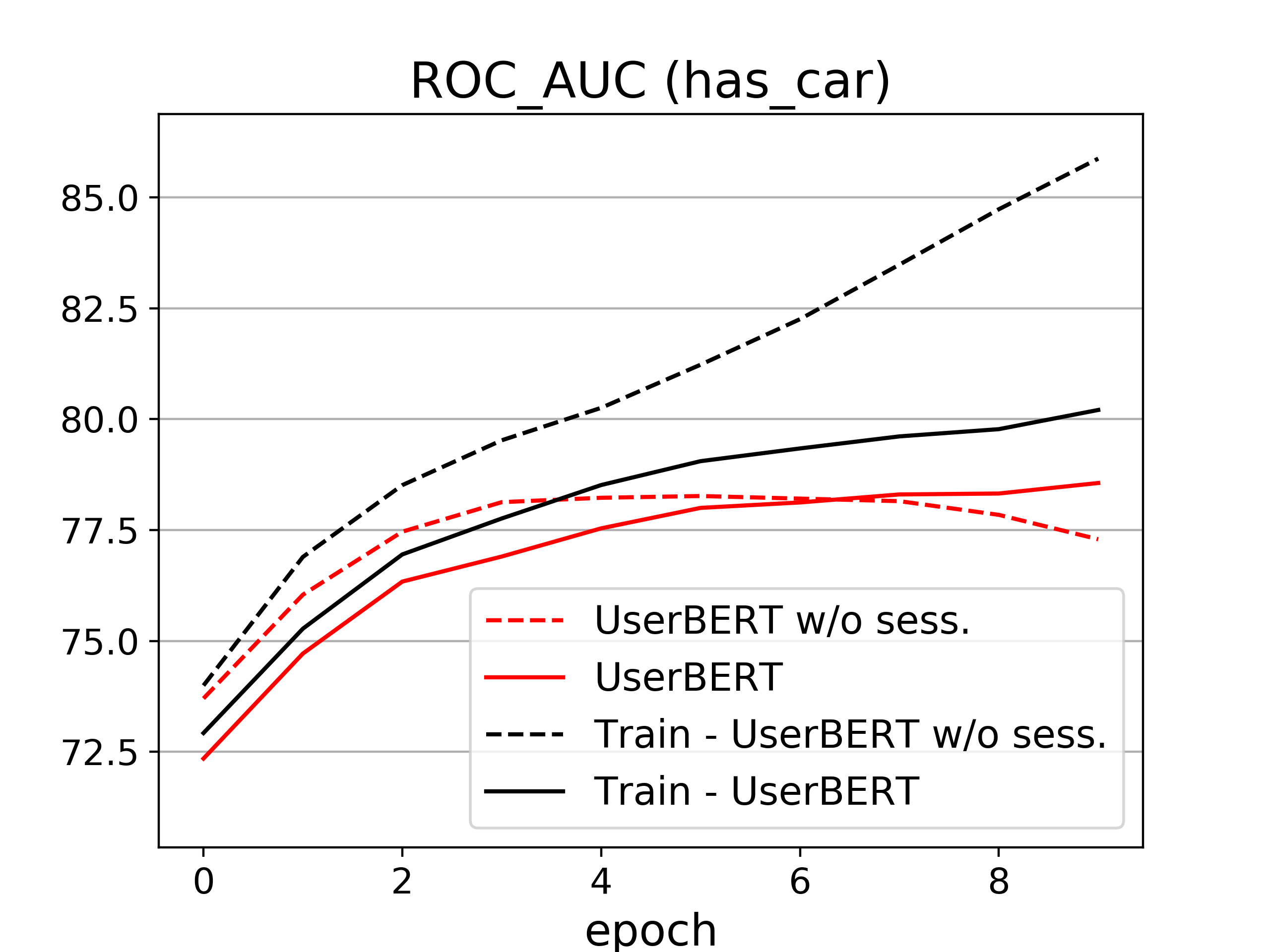}
    \end{subfigure}\hspace*{-1.05em}%
    \begin{subfigure}[b]{0.55\linewidth}
        \includegraphics[width=\linewidth]{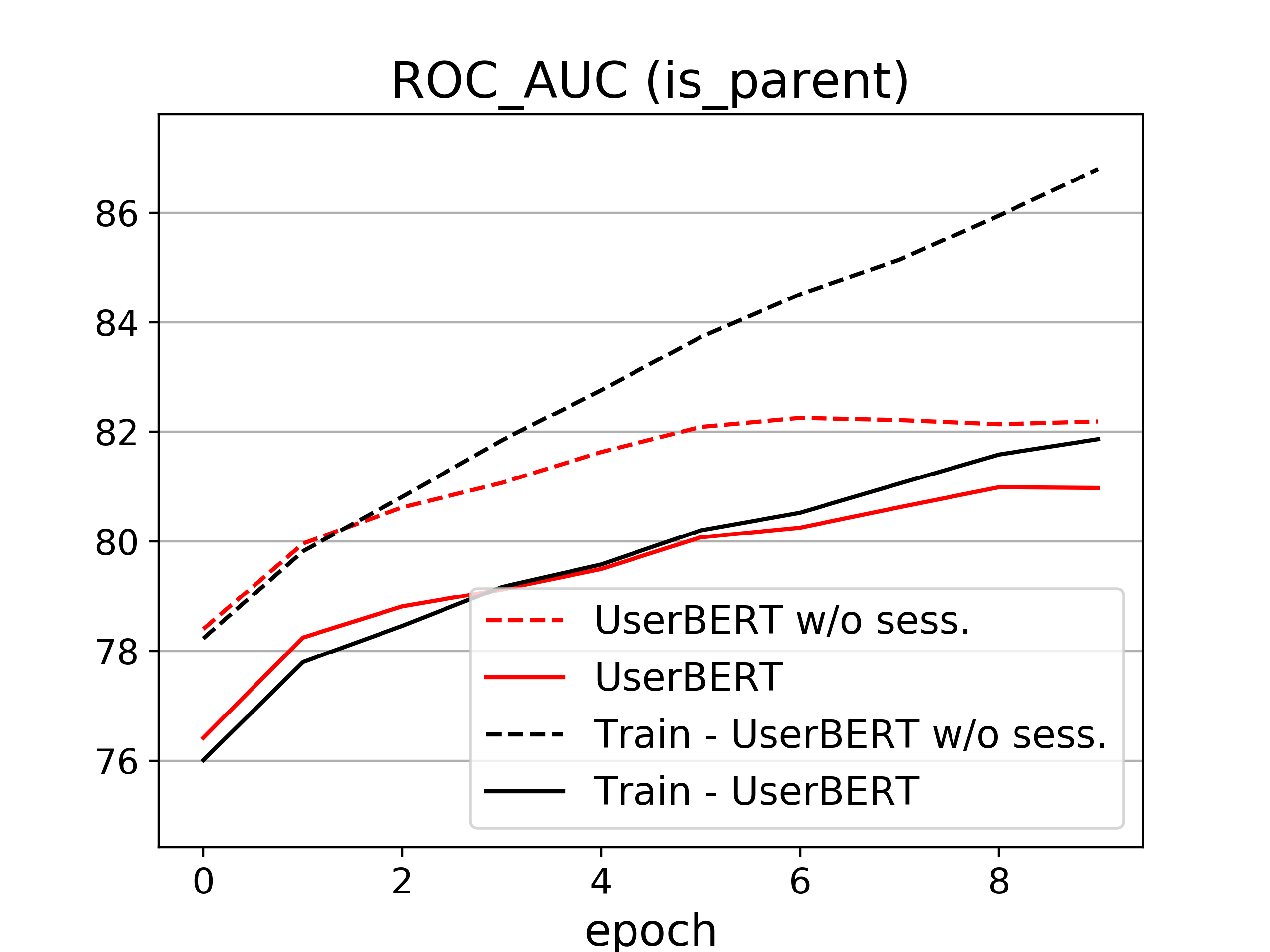}
    \end{subfigure}
    \caption{{\bf Effect of discretization of user actions into 'behavioral words'.} The plot shows fine-tuning performance results with and without discretization into sessions ('sess.') on \textit{user targeting} (above) and \textit{user attribute prediction} (below) tasks.}
    \label{fig:sessionization}
\end{figure}

\subsubsection{Effect of Additional Pretraining Labels}

We hypothesize that, compared to Transformer-based MTL, the learning of UserBERT is not limited by the multiple training tasks and is able to learn more expressive and generic representations from the input.

To further demonstrate the advantage of the proposed method over MTL-based pretraining, we pretrain Transformer-based MTL models with different numbers of labels before fine-tuning. We evaluate three training regimes: no pretraining, using 30\% of labels and using all labels. The comparison indicates that the performance of MTL is significantly affected by the number of training samples. As shown in Figure \ref{fig:user_targeting_fewer_labels}, more labeled data contributes to performance gains on the user targeting task. Models without the pretraining step show the worst performance.

In contrast, the pretraining of UserBERT does not require any additional collection of supervision signals, and therefore is not impacted by either the quantity or the quality of user annotations. 

\subsubsection{Effect of the Discretization of User Behavior}

In this study, instead of modeling every single user action, we discretize user action sequence into {\it behavioral words} for better generalization on downstream tasks.
Figure \ref{fig:sessionization} depicts the fine-tuning performance comparisons between pretrained models with and without the discretization of raw user action sequences for the user attribute prediction and user targeting tasks. 
In terms of ROC AUC comparison, the pretrained model with discretization improve fine-tuning performance on 2 of the 4 cases shown in the figure. 
Experimental results show that the discretization of user behavior improves next purchase genre prediction on mAP@10 by 2.1\%. 
In addition, the model without the discretization into behavioral words tends to overfit quickly as demonstrated in Figure~\ref{fig:sessionization}.

\section{Conclusion}
This paper introduces a new method to model user behavior by adapting the BERT model, which has made significant improvements in the NLP domain.
It explores and demonstrates the possibility for user-oriented machine learning tasks to alleviate the dependency on large annotated datasets.
We present UserBERT for pretraining user representations in a self-supervised manner on short-term and long-term behavior as well as user profile data.
We provide a novel method to tokenize raw user behavior sequences into behavioral words, which is demonstrated to reduce overfitting during pretraining.
Extensive experiments show that a well-designed pretrained model with self-supervision is able to outperform fully supervised learning models when transferred to downstream applications.

\bibliography{main}

\end{document}